\documentclass[conference]{IEEEtran}
\IEEEoverridecommandlockouts
\usepackage{cite}
\usepackage{amsmath,amssymb,amsfonts}
\usepackage{algorithmic}
\usepackage{graphicx}
\usepackage{textcomp}
\usepackage{subcaption}
\usepackage{dblfloatfix}
\usepackage{xcolor}
\def\BibTeX{{\rm B\kern-.05em{\sc i\kern-.025em b}\kern-.08em
    T\kern-.1667em\lower.7ex\hbox{E}\kern-.125emX}}
\begin{document}

\title{Leveraging BART to Assess CS1 C++ Programming Assignments using Rubric-based Criteria\\

}

\author{
\IEEEauthorblockN{Kelsey Rainey}
\IEEEauthorblockA{\textit{Department of Computer Science} \\
\textit{Tennessee Technological University}\\
Cookeville, TN \\
kkrainey42@tntech.edu}
\and
\IEEEauthorblockN{Jesse Roberts}
\IEEEauthorblockA{\textit{Department of Computer Science} \\
\textit{Tennessee Technological University}\\
Cookeville, TN \\
JTRoberts@tntech.edu}
}

\maketitle

\begin{abstract}

This paper investigates rubric-aware, multitask fine-tuning of transformer models for automated grading of introductory C++ programming assignments, with the goal of producing grade predictions that better reflect instructor grading behavior than general-purpose LLMs. Using multi-semester CS1 data, student submissions are paired with numeric scores, letter-grade buckets, and assignment rubrics, then preprocessed into unified sequences for transformer input. A BART encoder-decoder with LoRA adaptation is trained to jointly predict numeric grades and grade buckets, augmented with a distribution-matching term to align predicted and empirical grade distributions, an evaluation dimension often overlooked in prior work. Experiments compare single-task and multitask training, hard one-hot versus fuzzy and boundary-based soft labels, and rubric versus no-rubric conditions, with additional T5 and pairwise-pretrained variants. Results show that multitask BART with boundary-based soft labels and rubric context achieves lower mean absolute error and stronger grade-distribution alignment than single-task, hard-label, or code-only baselines. Fully fine-tuned T5 further improves distributional fidelity, while pairwise pretraining reduces numeric error at the cost of minority-class sensitivity. Collectively, the findings suggest that calibration-aware, rubric-guided training produces more instructor-like grading behavior than accuracy-optimized alternatives.
\end{abstract}


\section{Introduction}

Automated grading tools for introductory programming courses have the potential to reduce instructor workload, provide timely feedback to students, and scale assessment to larger class sizes. However, general-purpose large language models (LLMs) present a specific problem in this context: they are typically optimized for professional or advanced programming practices rather than the conventions taught in introductory courses, particularly in courses like the CS1 course examined here which uses C++. For example, introductory C++ courses commonly require students to use \texttt{using namespace std} to reduce syntactic complexity while core concepts are being learned, even though standard industry practice discourages this due to potential namespace conflicts. A model that penalizes such conventions, or rewards solutions that exceed the expected scope of the assignment, may grade in ways that conflict with instructor intent and course learning objectives.

This misalignment motivates the development of grading models fine-tuned on course-specific materials and rubrics. Beyond simply predicting a score, an effective automated grader should reflect the distribution of grades that an instructor would assign, distinguishing borderline submissions, avoiding collapse toward dominant grade categories, and handling the inherent subjectivity of rubric-based evaluation. These properties are not well captured by classification accuracy alone, particularly in introductory courses where grade distributions are often heavily skewed toward high marks.

This paper investigates how different training configurations affect the ability of transformer models to replicate instructor grading behavior on introductory C++ assignments. Starting from a BART encoder-decoder with LoRA adaptation as a base, we systematically examine the effect of multitask learning, labeling strategy, and rubric context on grading performance. We then extend the investigation to a T5-based model and a pairwise pretraining approach to assess whether these alternatives offer meaningful improvements. Model behavior is evaluated not only through standard metrics such as MAE, accuracy, and macro-F1, but also through Jensen-Shannon divergence between predicted and true grade distributions, a measure of whether a model reproduces realistic grading behavior rather than simply maximizing point accuracy.

Specifically, experiments address whether multitask learning outperforms single-task alternatives, whether soft labeling better captures grading uncertainty than hard one-hot encoding, and whether rubric context improves calibration. This work is intended as a foundation for a broader course-specific grading and feedback system, with the longer-term goal of supporting students who cannot immediately access instructional staff and need feedback aligned with their course's specific expectations rather than general programming conventions.

\section{Background and Related Work}

Artificial intelligence usage by educators and students alike has become increasingly prominent because of its ability to streamline tasks such as grading, generating feedback, explaining answers, or providing guided assistance. In introductory computer science courses, however, the impact of generative AI is difficult to measure because student experiences vary widely. Some novice programmers are able to effectively integrate AI into their workflow, while others struggle to learn both programming concepts and how to use AI tools effectively \cite{prather2024}.

In a study by Pankiewicz \cite{pankiewicz2024navigating}, students in an Object-Oriented Programming course were provided hints generated by ChatGPT. The hints are aimed to help novice students resolve compiler errors, runtime errors, or a failed unit test by explaining the error and expected solution. Results showed that 46\% of students found the hints useful, while 19\% reported that the hints were not helpful. Pankiewicz concluded that while the hints were useful, further research could be done into optimizing the hints to meet student needs and expectations. 

\subsection{AI in Code Grading}

Research on AI-assisted code grading has explored a variety of model architectures and evaluation approaches. One study compared BERT-based grading models against LSTM and TF-IDF approaches and found that BERT outperformed both alternatives because of its bidirectional encoding and stronger semantic understanding of code \cite{jun2025}. Other studies have identified additional limitations associated with using LLMs for code grading. Smaller models often produce unreliable feedback, and general-purpose LLMs can sometimes outperform code-specific models \cite{pan2024codev}. Subjective grading criteria, such as readability and coding style, may also introduce inconsistency and variance during training and evaluation \cite{zhang2024}. Additionally, another study reported that AI graders tend to assign harsher grades than human evaluators \cite{yousef2025}.

\subsection{AI in Rubric-Based Grading}

Rubric-based grading has been studied extensively in short-answer assessment and shares similarities with code grading since multiple solutions may satisfy the same learning objectives. Rubrics help models generalize across varied responses by emphasizing the relevant criteria for grading. Prior work has shown that rubric-based scoring can improve grading accuracy and consistency for short-answer responses \cite{senanayake2024}. Studies using LLMs for rubric grading report moderate to strong agreement with human evaluators, particularly when the rubric is clearly structured and detailed \cite{zhang2024}.

Research focused specifically on computer science grading has also shown that the method used to provide rubrics to the model affects performance. Supplying the complete rubric at once produced results that aligned more closely with human grading than evaluating one criterion at a time. Overall, rubric-based approaches consistently outperformed models that did not use rubrics \cite{pathak2025}.

While prior research has explored AI-assisted grading, rubric-based evaluation, and the use of transformer models for code assessment independently, fewer studies have examined how these approaches can be combined into a specialized multitask grading system for introductory computer science education. Existing work often evaluates general-purpose LLMs or focuses solely on either numeric scoring or categorical assessment, rather than training a model to perform both tasks simultaneously. Additionally, many studies rely on prompt engineering with pretrained models instead of fine-tuning models on course-specific assignments and instructor grading practices. This research differs by developing a fine-tuned, multitask, rubric-based LLM designed specifically for introductory programming courses. The model incorporates rubric context directly into the grading process while jointly predicting numeric grades and categorical performance levels, allowing shared learning between related grading tasks. By combining multitask learning with course-specific rubric guidance, this work investigates whether a specialized grading model can achieve stronger agreement with instructor evaluations than existing generalized AI grading approaches.

\subsection{Large Language Model}
This research employs BART \cite{DBLP:journals/corr/abs-1910-13461} and T5 \cite{raffel2020exploring} as its base architecture, a full encoder-decoder transformer that combines the bidirectional contextual encoding of BERT with the auto-regressive generative decoding of GPT. This architectural choice is significant from a theoretical standpoint. First, encoder-only models such as BERT are incapable of recursion and therefore limited in their computational expressivity \cite{roberts2024powerful}. Secondly, decoder-only models operate as language generators, whereas the full auto-encoder is more appropriate for sequence to sequence tasks. By retaining the full tranformer structure these limitations are avoided, making the models well-suited for tasks that require deep comprehension of an input sequence followed by structured prediction.

\section{Experiment Setup}

For each of the research questions, the same base model was trained on the same dataset of student code. This section details the data collection and pre-processing as well as the base model implementation. Each research question will further expand on how the base model was implemented to answer that question specifically.

\subsection{Data Cleaning and Pre-processing}

After collection, the data were standardized through a combination of lightweight manual checks and Python scripts. Submission archives were unzipped, nested directories flattened, and only the most recent submission per student and assignment was retained based on timestamped folder names. Semester-level grade sheets with heterogeneous headers were merged into a single file containing only student identifiers, assignment grades, and a semester label to link each submission to the correct rubric. Duplicate student identifiers across semesters (for repeat students) were disambiguated by appending a small suffix, and rubric PDFs were batch-converted to plain-text files using a consistent naming convention. The final dataset includes 2,404 samples with a grade distribution dominated by A-level submissions.

The main preprocessing pipeline focused on transforming code and rubrics into model-ready inputs. Cleaned records were loaded from CSV, rows missing key fields (semester, program number, numeric grade, grade bucket, or code text) were removed, submissions without a matching rubric were filtered out, and grade buckets were restricted to A–F. Numeric grades were clamped and normalized to the 0–100 range to stabilize regression and facilitate comparison across assignments. 

For the model input, C++ submissions were normalized and combined with rubric text into a single sequence. 
In the dataset class, each example loaded the corresponding rubric text from disk, concatenated it with the preprocessed code using section markers \texttt{[RUBRIC]} and \texttt{[CODE]}, and passed the resulting sequence through a BART tokenizer with truncation and fixed-length padding up to \texttt{MAX\_LENGTH} tokens. Each instance stored a normalized numeric grade, a discrete bucket ID, and optionally a fuzzy membership vector derived from the numeric grade for soft-label experiments.

\subsection{Base Model}

The model used is a pretrained BART encoder–decoder augmented with low-rank adaptation (LoRA) \cite{hu2022lora} on the query and value projections of the attention layers \cite{DBLP:journals/corr/abs-1910-13461}, using rank \(r=32\), scaling factor \(\alpha=64\), and zero LoRA dropout, while only the adapters, task-specific heads, and final encoder block are updated. Across experiments, the BART encoder processes rubric text and the decoder processes the corresponding student code; decoder hidden states are pooled into a joint rubric–code representation that feeds a regression head for numeric grades and a classifier over five buckets (A–F), trained with a multitask loss combining mean-squared error on normalized grades, a mixture of focal and soft-label cross-entropy on fuzzy bucket memberships, and a KL regularizer that aligns the batch-averaged predicted distribution with the empirical grade distribution. Optimization uses AdamW with learning rate \(5\times10^{-5}\), zero weight decay, batch size 2, gradient clipping at norm 1.0, a linear decay schedule without warm-up over 20 epochs, and a weighted sampler that upsamples underrepresented grade buckets, selecting the checkpoint with the best validation macro-F\(_1\) for reporting.

\section{RQ 1: Does a multitask model outperform the single-task counterparts?}
To answer this question, we compared multitask and single-task variants on accuracy and MAE as well as how closely their predicted bucket distributions match the empirical distribution of grades based on Jensen-Shannon Distance. Prior work in multitask learning suggests that jointly learning related tasks can improve difficult prediction problems by leveraging shared representations and additional training signals from related \cite{caruana1997multitask}. Building on this idea, we evaluate whether jointly predicting numeric grades and grade buckets, together with the distribution-matching loss, produces outputs that better reflect the grading patterns present in the data.

\subsection{Methodology}

To setup, two task-specific heads are applied to the base model: a regression head and a classification head.  The regression head is a two-layer perceptron with sigmoid activation and dropout that outputs a single scalar to obtain a numeric grade prediction. The classification head has the same structure but outputs a 5‑dimensional logit vector over the grade buckets A–F. Training looks to minimize a weighted sum of mean-squared error on the normalized numeric grade and a focal classification loss on the bucket logits.

At evaluation for the multitask model, a constrained bucket prediction is also provided that enforces consistency between the predicted numeric grade and discrete bucket. For each example, the predicted numeric grade is first converted into a small set of plausible buckets, then mask out all other class logits and re-normalize by selecting the highest logit within the allowed set.

Depending on the task, the model had different assessment criteria. For the regression task, mean absolute error (MAE) was used to determine how far the predicted point values were from the true grades on average. For the classification task, several additional evaluation metrics were reported. The first was the confusion matrix, which helped visualize how the model predicted grade labels compared to the true grade labels. This was important for determining whether the model was making predictions across all grade buckets rather than collapsing into predicting a single dominant grade. For the classification task specifically, matching the overall distribution of grades mattered more than raw accuracy alone. The original dataset is heavily skewed toward \textit{A} grades, meaning a model that predicts only \textit{A} could still achieve deceptively high accuracy while failing to learn the nuances of evaluation. To better measure how closely the predicted grade distribution aligned with the true distribution, Jensen-Shannon Divergence (JSD) was also used, where lower divergence values indicate that the predicted distribution more closely matches the original grading distribution. Accuracy and macro-F1 score were additionally reported for comparison, though they served as secondary evaluation measures relative to distribution alignment. For the multitask model, all regression, classification, and distribution-based metrics were reported together.

\subsection{Results}
The multitask BART model achieved the strongest overall performance across both regression and classification objectives. As shown in Table \ref{tab:task_comparison}, the multitask model produced the lowest validation MAE (11.97), indicating the closest alignment with instructor-assigned numeric grades, while also achieving the highest accuracy (56.1\%). Although the constrained multitask model achieved a slightly higher macro-F1 score (0.327 versus 0.326), its overall accuracy and distribution alignment were lower than the unconstrained multitask model. The classification-only model and regression-only models performed worse across all reported metrics, suggesting that learning one task in isolation was less effective than jointly learning grading tasks together. Additionally, the multitask models achieved the lowest JSD values, indicating that multitask learning better preserved the original grade distribution and avoided collapsing toward dominant grade categories. Overall, these results suggest that combining regression and classification objectives in a multitask framework improves both numeric grading agreement and the model’s ability to reflect realistic grading distributions.

\begin{table}[ht]
\centering
\resizebox{\columnwidth}{!}{%
\setlength{\tabcolsep}{1pt}
\begin{tabular}{lcccc}
\hline
Model & Val MAE & Macro-F1 & Accuracy & JSD \\
\hline
BART Multitask            & \textbf{11.97} & 0.326 & \textbf{0.561} & \textbf{0.007055} \\
BART Multitask Constrained&  N/A  & \textbf{0.327} & 0.524 & 0.009990 \\
BART classification-only  &  N/A  & 0.297 & 0.472 & 0.009696 \\
BART regression-only      & 12.21 &  N/A  &  N/A  &  N/A  \\
\hline
\end{tabular}%
}
\caption{Evaluating Multi-task vs single task performance}
\label{tab:task_comparison}
\end{table}

\begin{figure}[h]
    \centering
    \includegraphics[width=\columnwidth]{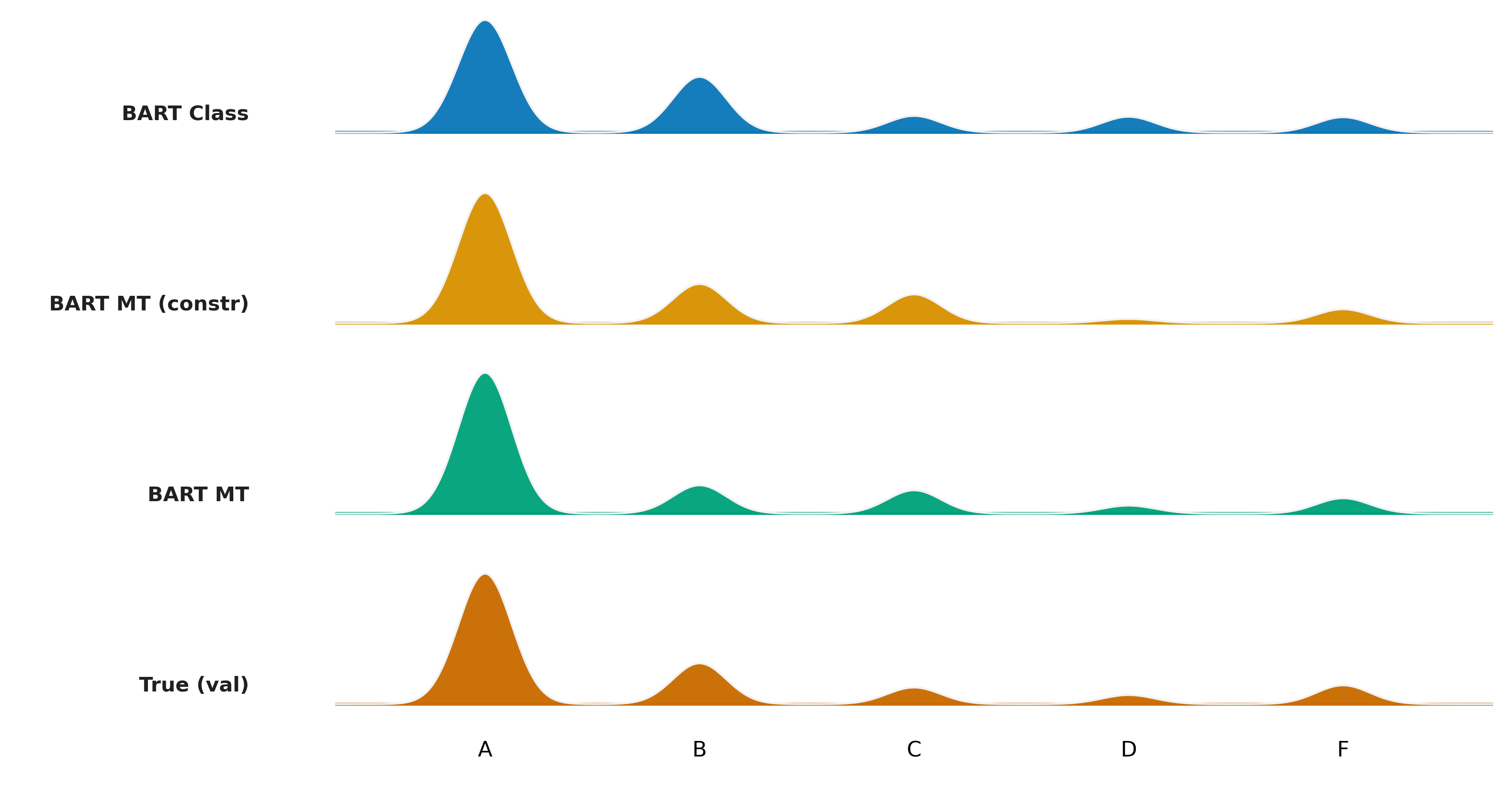}
    \caption{Distribution comparison of models for RQ1}
    \label{fig:RQ1_Dist}
\end{figure}

\section{RQ2: Does hard vs soft labels affect model predictions?}
The next research question was asked to determine if soft labels could better capture the data distribution compared to the traditional one-hot encoding. Classification tasks typically use one-hot encoding to turn each potential category into machine-readable numbers. In this case, there would be five numbers that represent the potential classification of a grade, with the position indicating which number corresponds to which class. A grade belonging to the the \textit{B} class would would be denoted by 01000, etc. In an attempt to capture the nuance of multiple graders or variation among similar assignments, Fuzzy membership was also used as a comparison to one-hot encoding. Fuzzy membership allows for each potential class to have a degree of belonging \cite{mamdani1977application}. For example, a 90 can be represented as 0.6 \textit{A}, 0.4 \textit{B}, 0 \textit{C}, 0 \textit{D}, and 0 \textit{F} showing that while it is mostly an A, there is some degree that it belongs to \textit{B}. The hypothesis being that the soft labels may be better able capture the grading as it mirrors how different human graders may approach grading one assignment. 

\subsection{Methodology}

To test this, the base BART script was changed to include a fuzzy membership as well as soft cross-entropy as the base loss function. A boolean was added to the script as well to be able to more easily switch between the fuzzy membership setup and one-hot encoding. Two different implementations for fuzzy membership were tested: a full triangular fuzzy membership and fuzzy membership only near the boundary grades. 

The fuzzy\_membership\_from\_numeric function is the full fuzzy membership implementation. The function takes a numeric grade and normalizes it. Then, the function uses overlapping triangular membership functions centered at the mid-point for the \textit{A}/\textit{B}/\textit{C}/\textit{D}/\textit{F} scores. For example, \textit{A} is centered at 95, \textit{B} at 85, and so on. The half-widths are used to determine the width of each triangle. The function then computes the raw membership values and ensures that every grade is active in at least one bucket, defaulting to As for high grades and Fs for low grades. Finally, the function normalizes the vectors so the membership across all buckets sums to 1. The figure below shows the representation of each grade bucket given the full fuzzy membership function. In order to get a better representation of \textit{D}, which was the lowest-represented class, the midpoint of \textit{F} was moved down to give \textit{D} a larger width of D-only grades. 

\begin{figure}[h]
    \centering
    \includegraphics[width=\columnwidth]{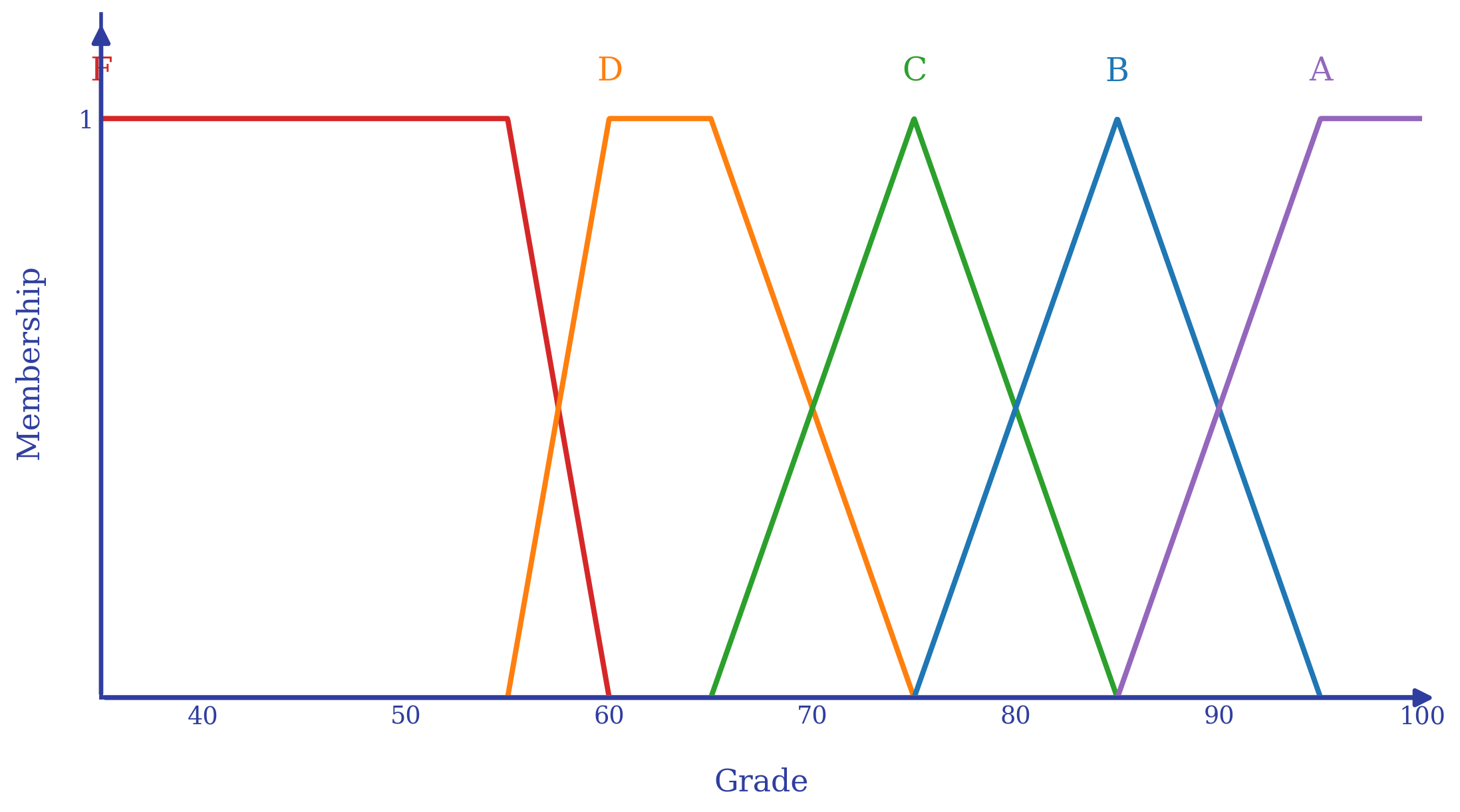}
    \caption{Visual of full fuzzy membership buckets}
    \label{fig:my_image}
\end{figure}

Alternatively, the boundary\_soft\_label function takes a numeric grade and existing “hard” letter bucket. The function defines numeric boundaries between consecutive buckets and looks for the closest boundary to the grade. If the grade lies within a specified window around the boundary, it smoothly interpolates between the two neighboring buckets to assign a soft label shared between them. If the grade does not lie within the boundary window, the function keeps a one-hot label on the provided hard bucket. The figure below shows the bucket representation with the soft labels only at the boundaries of each grade bucket.

\begin{figure}[h]
    \centering
    \includegraphics[width=\columnwidth]{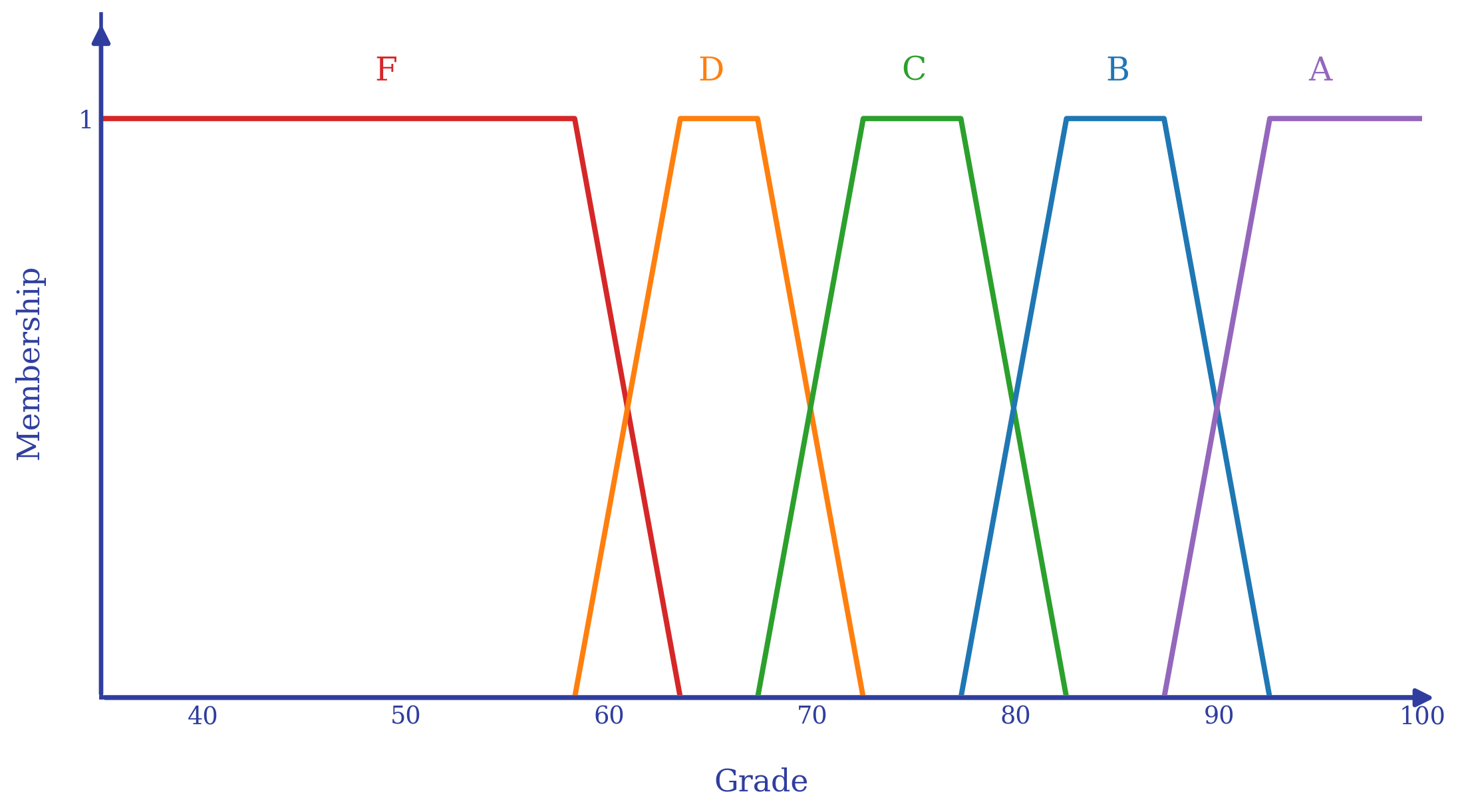}
    \caption{Visual of boundary fuzzy membership buckets}
    \label{fig:my_image}
\end{figure}

Both versions of the soft labeling scripts as well as a hard labeling script were run, keeping as much similar between the versions as possible to ensure that the differences could mainly be attributed to the labeling and not differences in the model setup. For fuzzy membership training runs, a focal classification loss on the bucket logits is interpolated with a soft-label cross-entropy term.

\subsection{Results}
The results laid out in Table \ref{tab:label_comparison} show that the soft labels improve the regression alignment, with the full fuzzy membership having the lowest MAE. The one-hot encoding has the best accuracy and macro-F1 score. However, the distribution shows us that the one-hot encoding over predicts A, which can explain the high accuracy because the dataset is heavily skewed towards A. The fuzzy at the boundary implementation has the lowest JSD, indicating the closest predicted distribution to actual distribution with the one-hot distribution following. 


\begin{table}[ht]
\centering
\resizebox{\columnwidth}{!}{%
\setlength{\tabcolsep}{1pt}
\begin{tabular}{lcccc}\hline
Model & Val MAE & Macro-F1 & Accuracy & JSD \\
\hline
BART full fuzzy             & \textbf{11.29} & 0.308 & 0.457 & 0.018333 \\
BART full fuzzy (constr)    &  N/A  & 0.306 & 0.453 & 0.040349 \\
BART fuzzy boundary         & 11.83 & 0.306 & 0.483 & \textbf{0.002314} \\
BART fuzzy boundary (constr)&  N/A  & 0.316 & 0.482 & 0.013386 \\
BART one-hot                & 11.97 & 0.326 & \textbf{0.561} & 0.007055 \\
BART one-hot (constr)       &  N/A  & \textbf{0.327} & 0.524 & 0.009990 \\
\hline
\end{tabular}%
}
\caption{Evaluating fuzzy vs crisp encoding}
\label{tab:label_comparison}
\end{table}

\begin{figure}[h]
    \centering
    \includegraphics[width=\columnwidth]{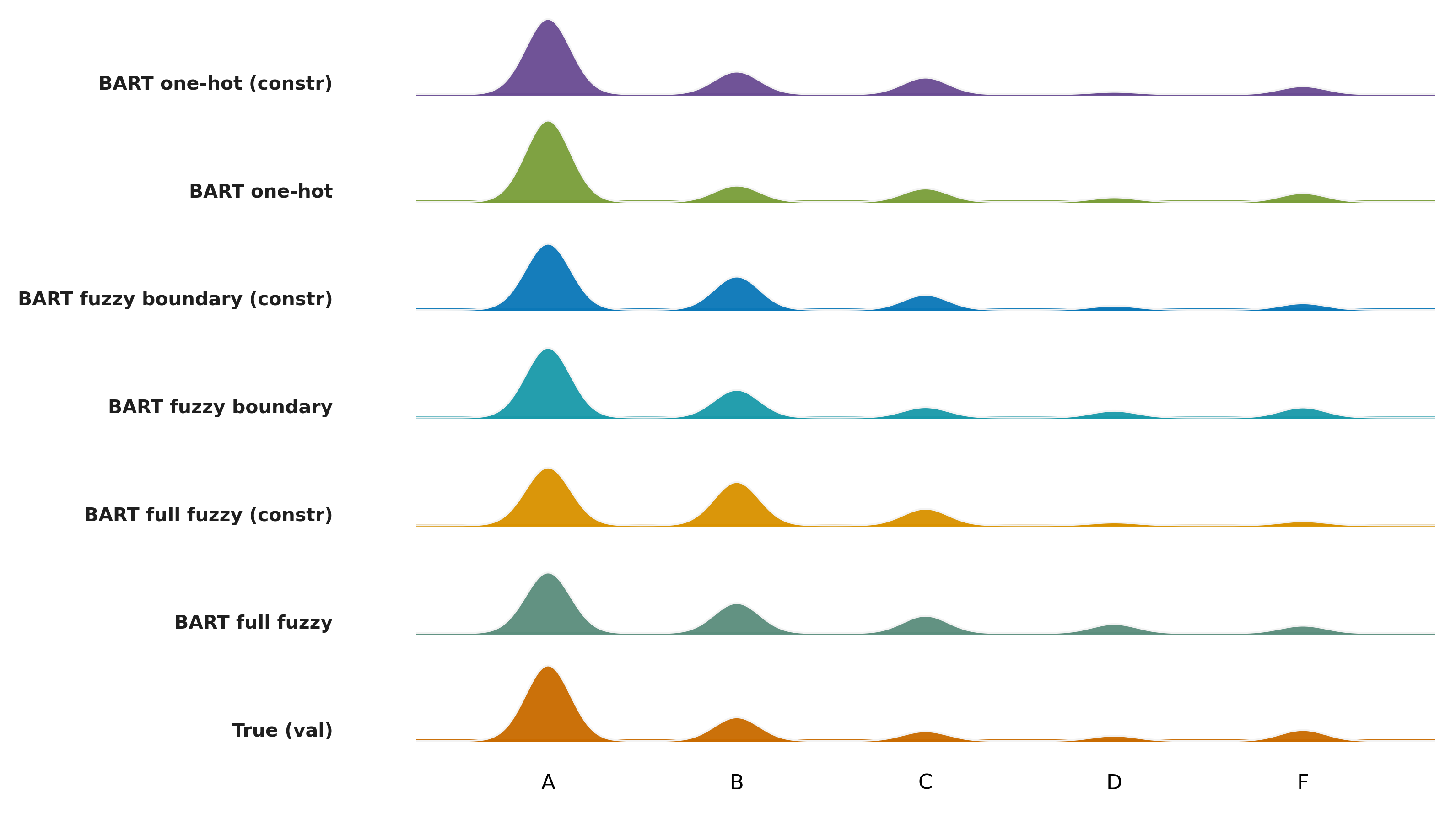}
    \caption{Distribution comparison of models for RQ2}
    \label{fig:RQ2_Dist}
\end{figure}

The predicted grade distributions in Figure \ref{fig:RQ2_Dist} illustrate these results more clearly. The one-hot models produce distributions dominated by \textit{A} predictions, with noticeably reduced representation of lower grades. This behavior explains the stronger accuracy results, since predicting the majority class more aggressively increases overall correctness under class imbalance. However, it also leads to less balanced grade predictions. The constrained one-hot model slightly reduces this effect, but still remains strongly concentrated around \textit{A} grades.

By contrast, the full fuzzy membership models generate much smoother distributions. Rather than sharply separating grades, these models redistribute probability mass toward neighboring categories, especially the middle grades of \textit{B} and \textit{C}. This smoothing behavior likely explains the lower MAE values: predictions that fall between categories reduce the magnitude of numeric error even when the exact class label is incorrect. However, the constrained full fuzzy model appears to overcorrect this behavior. The distribution plot shows that many low-grade predictions that would otherwise fall into \textit{D} or \textit{F} are instead shifted upward into \textit{B} and \textit{C} categories. As a result, minority classes become severely underrepresented, increasing the JSD despite competitive MAE performance.

The fuzzy boundary approach provides a more balanced compromise between these extremes. Because only samples near grade cutoffs receive softened labels, the model preserves clearer category structure while still allowing flexibility near uncertain boundaries. In the distribution plots, the fuzzy boundary models retain the overall shape of the true distribution more closely than either the full fuzzy or one-hot approaches. This behavior is reflected quantitatively by the lowest JSD value among all models. At the same time, the fuzzy boundary models maintain competitive MAE, accuracy, and macro-F1 scores, suggesting that boundary-based softening improves calibration without collapsing predictions toward either the majority class or the middle-grade categories.

Overall, the constrained variants do not appear to consistently improve performance. Instead, the constraint tends to compress predictions toward the center of the grading scale, particularly for the fuzzy-label approaches. This effect reduces the frequency of extreme predictions such as \textit{D} and \textit{F} grades and weakens alignment with the true class distribution. While the constraints may encourage consistency between numeric and categorical outputs, the results suggest that they do so at the cost of realistic grade distributions and minority-class representation.

\section{RQ 3: Does the rubric as context improve model predictions?}
This experiment tests whether exposing the model to the instructor’s stated expectations leads to more calibrated, distribution‑matching, and semantically faithful grades—for example, better distinguishing borderline cases according to the rubric rather than just surface code patterns. Comparing rubric and no‑rubric models under the same training and evaluation pipeline can determine whether the rubric improves accuracy and calibration, helps the predicted bucket distribution align with the empirical distribution, or mainly acts as noise when rubrics are inconsistent or redundant with the code.

\subsection{Methodology}
The rubric condition followed the standard architecture where the encoder  received the rubric text and the decoder processed the student code. To determine if the rubric context was helpful to the model, a no-rubric variant was made where the model only sees the code and an empty rubric as input, so all patterns learned are from code structure and label correlations. For the base model, the most successful setup determined by RQ1 and RQ2 was used (fuzzy boundary) for both the rubric and no-rubric models. 

\subsection{Results}
The rubric comparison results in Table \ref{tab:rubric_comparison} show that incorporating rubric context generally improved numeric grading performance and distribution alignment compared to models trained without rubric information. The best overall regression performance came from the fuzzy boundary rubric model, which achieved the lowest validation MAE (11.83) and the lowest JSD (0.002314), indicating the closest agreement with instructor-assigned numeric grades and the strongest match to the true grade distribution. The standard one-hot rubric multitask model also performed strongly, achieving the highest overall accuracy (56.1\%) while maintaining a low MAE (11.97), suggesting that rubric context helped improve both grading precision and categorical prediction consistency.

The constrained fuzzy boundary model achieved the highest macro-F1 score (0.316), indicating slightly better balanced classification performance across grade categories, though this came with worse distribution alignment and no regression evaluation. In contrast, the models trained without rubric information consistently produced higher MAE values and worse JSD scores. 

\begin{table}[ht]
\centering
\resizebox{\columnwidth}{!}{%
\setlength{\tabcolsep}{1pt}
\begin{tabular}{lcccc}
\hline
Model & Val MAE & Macro-F1 & Accuracy & JSD \\
\hline
BART no rubric fuzzy        & 12.67 & 0.294 & 0.455 & 0.018569 \\
BART fuzzy boundary         & \textbf{11.83} & 0.306 & 0.483 & \textbf{0.002314} \\
BART no rubric one-hot      & 13.18 & \textbf{0.339} & 0.534 & 0.005975 \\
BART one-hot                & 11.97 & 0.326 & \textbf{0.561} & 0.007055 \\
\hline
\end{tabular}%
}
\caption{Evaluating the impact of rubric inclusion}
\label{tab:rubric_comparison}
\end{table}

\begin{figure}[h]
    \centering
    \includegraphics[width=\columnwidth]{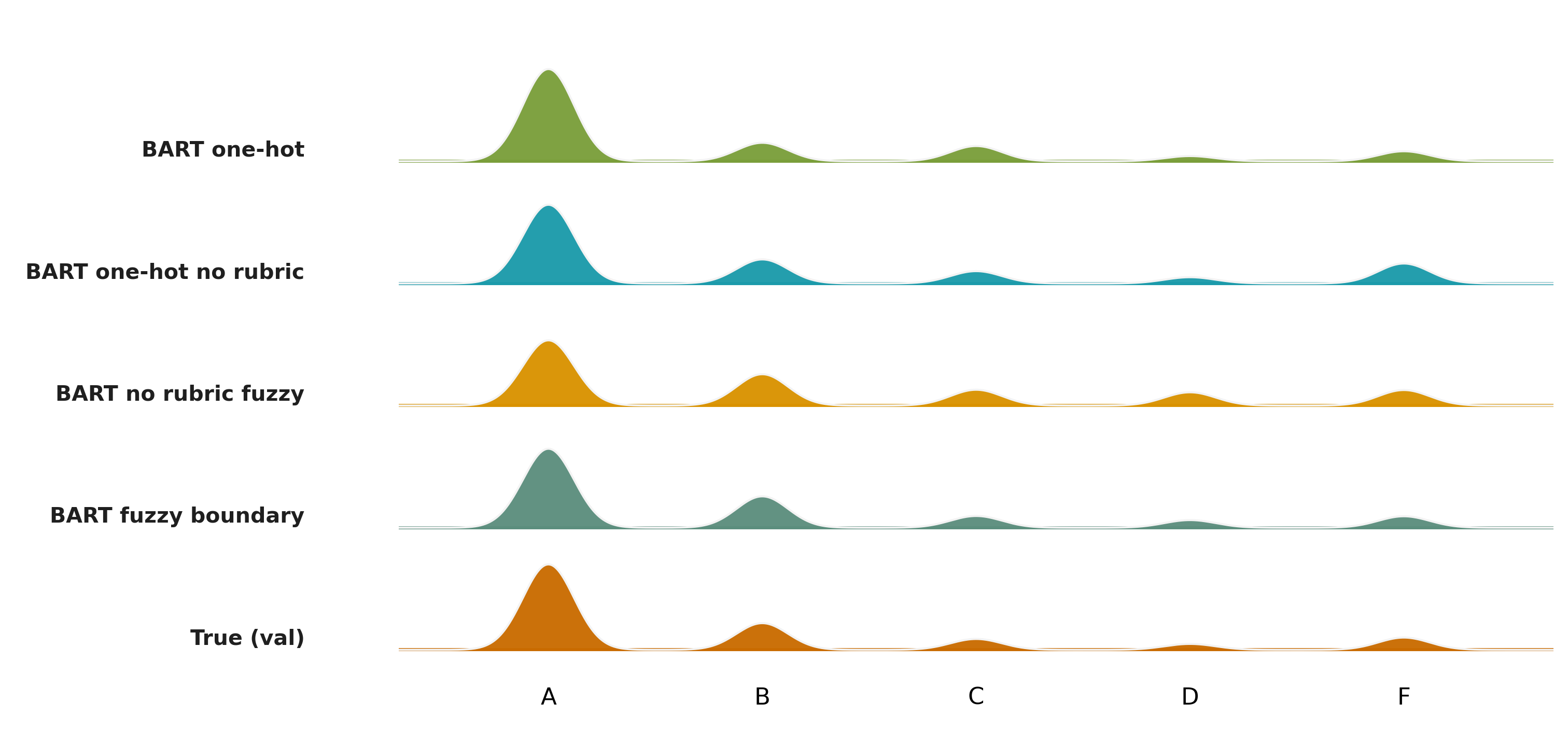}
    \caption{Distribution comparison of models for RQ3}
    \label{fig:RQ3_Dist}
\end{figure}

The distribution plots in Figure \ref{fig:RQ3_Dist} help explain these differences. The no-rubric models exhibit noticeably less stable grade distributions, particularly in the lower-performing categories. The no-rubric one-hot model still strongly favors \textit{A} predictions, but also produces a disproportionately large number of \textit{F} predictions compared to the true distribution. This suggests that without rubric guidance, the model relies more heavily on coarse class separation and becomes less calibrated in distinguishing borderline submissions. While this sharper separation contributes to the highest macro-F1 score among the compared models, it also increases numeric grading error, resulting in the worst MAE overall.

The rubric-enhanced one-hot model moderates this behavior. Although it still overpredicts \textit{A} grades relative to the true distribution, the addition of rubric context reduces the exaggerated F-category predictions and improves both MAE and overall accuracy. This suggests that rubric information helps differentiate between nearby performance levels rather than collapsing uncertain examples into extreme categories.

The fuzzy-label models show a different pattern. The no-rubric fuzzy model produces smoother predictions than the one-hot approaches, but its distribution remains less aligned with the true validation set, particularly in the middle-grade categories. Once rubric context is added through the fuzzy boundary approach, however, the predicted distribution becomes substantially closer to the ground truth. The fuzzy boundary rubric model preserves the overall shape of the real distribution while still maintaining competitive classification metrics. Compared to the one-hot models, it avoids both the heavy concentration of \textit{A} grades and the exaggerated tail behavior in the \textit{F} category.

These results suggest that rubric information improves calibration rather than simply improving classification accuracy. The rubric-based models demonstrate more realistic prediction distributions and lower numeric grading error.

\section{Impact of Model Size and Pretraining}

Here we investigate the impact of model size and the potential impact of an added pretraining to improve the representation learned by the model. 

\subsection{T5}

We use a T5 model based on the BART Multitask findings with fuzzy boundary function. T5 was compared using a full fine-tuning as well as a LoRA fine-tuned version to the results obtained from BART. 

\begin{table}[h]
\centering
\resizebox{\columnwidth}{!}{%
\begin{tabular}{lccc}
\hline
Metric & T5 Full & T5 LoRA & BART MT \\
\hline
Validation Accuracy  &   \textbf{0.514}  &  0.459   &   0.483  \\
Validation Macro-F1  &   0.331  &  \textbf{0.338}   &   0.306  \\
Validation MAE       &   12.01  &  12.37   &   \textbf{11.83}  \\
JSD                  & \textbf{0.002192} & 0.027201 & 0.002314 \\
\hline
\end{tabular}%
}
\caption{T5 and BART Comparison}
\label{tab:t5_comparison}
\end{table}

As shown in Table \ref{tab:t5_comparison}, the fully fine-tuned T5 model obtained the highest validation accuracy at 0.514 and achieved a strong macro-F1 score of 0.331, indicating improved balanced performance across grade categories. Most notably, the T5 Full model achieved the lowest JSD value (0.002192), slightly outperforming BART multitask (0.002314). This suggests that the T5 Full predictions most closely matched the true grade distribution, indicating strong calibration and balanced prediction behavior across classes.

The T5 LoRA model demonstrated competitive classification behavior despite using parameter-efficient fine-tuning. Although its overall accuracy was lower than both other models, it achieved the highest Macro-F1 score (0.338). This suggests that the LoRA approach may better preserve minority-class sensitivity or reduce bias toward majority grade categories. However, the substantially higher JSD value (0.027201) indicates that its predicted distribution deviated more significantly from the true grade distribution, implying poorer calibration and a tendency to over- or under-predict certain classes.

These results suggest that model size and fine-tuning strategy affects different aspects of grading performance. Full fine-tuning with T5 appears to improve classification accuracy and distributional fidelity, while BART multitask retains an advantage for continuous score prediction. The LoRA-based T5 implementation demonstrates that parameter-efficient tuning can remain competitive in macro-F1 performance, but may sacrifice calibration stability and overall accuracy. Collectively, the findings indicate that multitask learning combined with fuzzy boundary grading generalizes effectively across transformer architectures, while also highlighting trade-offs between efficiency, calibration, and predictive precision.

\subsection{Pairwise Pretraining}

The BART pairwise model was based on the BART Multitask with boundary fuzzy membership. A pretraining pairwise preference task was investigated in which the model was required to classify one of two submissions as superior.  

\begin{table}[h]
\centering
\setlength{\tabcolsep}{4pt}
\resizebox{\columnwidth}{!}{%
\begin{tabular}{lcc}
\hline
Metric & BART Pairwise & BART Multitask \\
\hline
Validation Accuracy &  \textbf{0.497}   &  0.483   \\
Validation Macro-F1 &  0.262   &  \textbf{0.306}   \\
Validation MAE      &  \textbf{11.66}   &  11.83   \\
JSD                 & 0.006787 &  \textbf{0.002314}\\
\hline
\end{tabular}%
}
\caption{Evaluating the impact of Pairwise Pretraining}
\label{tab:pairwise_comparison}
\end{table}

The pairwise task demonstrates an interesting tradeoff between ordinal accuracy and class-level discrimination. Compared to the multitask boundary model shown in Table \ref{tab:pairwise_comparison}, the pairwise model achieved lower MAE (11.66 vs. 11.83) and slightly higher accuracy (0.497), while maintaining a relatively low JSD (0.006787). This suggests that the pairwise objective improved the model’s ability to preserve the relative ordering of grades, reducing the magnitude of prediction errors, without substantially distorting the overall predicted grade distribution.

In contrast, the full fuzzy model (shown in Table \ref{tab:label_comparison}) achieved the best MAE overall (11.29), but at the cost of a significantly worse JSD (0.018333). Although the fuzzy labels encouraged smoother predictions, the model appeared to overfit toward certain grade regions, leading to poorer calibration of the predicted distribution. This indicates that lower MAE alone does not necessarily correspond to better distributional alignment.

The pairwise pretraining improves accuracy and MAE over the multitask model while avoiding the larger distributional shift observed in the full fuzzy approach. However, this came with a reduction in macro-F1 (0.262), suggesting that the model became less effective at distinguishing minority or boundary classes. In other words, the pairwise objective may encourage conservative predictions that stay closer to neighboring grade categories, improving average error magnitude while reducing sensitivity to less frequent classes.

\section{Discussion}

Grading performance cannot be evaluated effectively using accuracy alone, especially in highly imbalanced educational datasets. Across nearly all experiments, models achieving the highest accuracy also tended to overpredict the dominant \textit{A} category. This demonstrates that traditional classification metrics can obscure undesirable grading behavior when most students receive high grades. The inclusion of JSD and distribution analysis shows that a model may appear strong numerically while still failing to reproduce realistic grading behavior. More generally, this suggests that educational AI systems should be evaluated not only on point prediction accuracy, but also on calibration and distributional fidelity.

The findings also highlight a tradeoff between sharp decision boundaries and grading realism. Models optimized primarily for classification metrics favored aggressive majority-class predictions, while models that incorporated uncertainty and contextual information produced smoother, more calibrated outputs. This implies that the “best” grading model depends on the intended educational goal. If the objective is maximizing exact categorical agreement, harder labels and sharper boundaries may be preferred. If the objective is producing instructor-like grading behavior that reflects uncertainty and preserves realistic grade distributions, softer boundary-aware approaches appear more appropriate.

More generally, the results suggest that automated grading systems for introductory programming may benefit from being designed as calibrated evaluators rather than pure classifiers. Human grading is often subjective, boundary-sensitive, and influenced by rubric interpretation. Models that incorporate multitask learning, rubric context, and limited soft-label uncertainty appear better able to approximate these characteristics than models trained purely for categorical accuracy.

\section{Limitations and Future Work}
One limitation of this study was the heavily skewed grade distribution. Most submissions received \textit{A} grades, which led several models to collapse toward predicting only \textit{A} outcomes. A more balanced dataset would likely improve the model’s ability to distinguish between performance levels and generalize across grade categories.

Another limitation is the dataset had to be assembled from multiple sources, including rubric PDFs, grade spreadsheets, and archived student submission files. In many cases, these components were stored separately and inconsistently across semesters, requiring additional effort to ensure they could be reliably linked together. A key challenge was aligning rubrics and grades with the correct student submissions, since neither was originally designed for direct machine learning use. As a result, substantial preprocessing was required to standardize file formats, establish consistent identifiers, and ensure that each submission could be correctly paired with its corresponding rubric and grade entry. This also limited the amount of data that could be used as not all semesters had all three components available for use.

Future work will focus on adding a feedback generation component that produces rubric-based comments alongside predicted grades. Another direction for experimentation is incorporating generated feedback into the grading process itself, allowing the model to use a completed rubric or feedback context when predicting final grades. The research questions as well could be incorporated into the extensions, such as exploring the rubric and no-rubric implementations or the labeling approach for the T5 and pairwise models. 

\section{Conclusion}

This research demonstrates that transformer-based grading systems can be adapted to better reflect instructor grading in introductory programming courses when trained with course-specific context and evaluation objectives. Rather than relying solely on traditional classification accuracy, the results show that grading systems benefit from approaches that account for calibration, uncertainty, and realistic grade distributions.

The findings suggest that different model configurations are appropriate depending on the intended grading goal. Models trained with one-hot labels achieved the strongest overall accuracy, making them more suitable when maximizing exact categorical agreement is the primary objective. In contrast, soft-label approaches, particularly the fuzzy boundary implementation, produced grade distributions that more closely aligned with instructor grading behavior while maintaining competitive performance across other evaluation metrics. This indicates that introducing limited uncertainty near grade boundaries may better capture the subjectivity present in human grading.

Across nearly all experiments, multitask learning and rubric integration consistently produced the strongest performance. Jointly learning numeric and categorical grading tasks improved both regression and classification, while rubric context helped the models better distinguish between nearby performance levels rather than relying only on superficial code patterns. Together, these findings suggest that automated grading systems for introductory programming should be designed as context-aware evaluators rather than simple classifiers.

The extension experiments also highlight that architecture and fine-tuning decisions involve practical tradeoffs rather than universally superior choices. Full fine-tuning with larger models such as T5 improved distribution alignment and balanced classification behavior, while LoRA-based approaches remained competitive despite significantly lower parameter costs. Similarly, pairwise pretraining improved ordinal grading behavior and reduced average numeric error, though at the expense of minority-class sensitivity. These results suggest that model selection may depend as much on available computational resources and deployment constraints as on small differences in evaluation metrics.

One major challenge throughout this work was the heavily imbalanced dataset, where most submissions received high grades. This imbalance encouraged several models to collapse toward predicting dominant categories, particularly \textit{A} grades. A more evenly distributed dataset would likely reduce the need for distribution-matching techniques and allow the models to better learn distinctions between lower-performing submissions. Future work can further explore balancing strategies, feedback generation, and the integration of rubric-based explanations directly into the grading process.

Overall, this work provides a foundation for AI-assisted grading systems tailored to introductory programming education. The results demonstrate that incorporating multitask learning, rubric guidance, and uncertainty-aware labeling can produce grading behavior that more closely resembles human evaluation while remaining flexible enough to adapt to different instructional priorities and resource constraints.

\bibliographystyle{IEEEtran}
\bibliography{Ref.bib}

\vspace{12pt}
\color{red}

\end{document}